\documentclass[10pt,twocolumn,letterpaper]{article}

\usepackage{cvpr}      

\usepackage{graphicx}
\usepackage{amsmath}
\usepackage{amssymb}
\usepackage{booktabs}
\usepackage[noend]{algpseudocode}

\usepackage{algorithmicx,algorithm}

\usepackage[pagebackref,breaklinks,colorlinks]{hyperref}
\usepackage{multirow}

\usepackage[capitalize]{cleveref}
\crefname{section}{Sec.}{Secs.}
\Crefname{section}{Section}{Sections}
\Crefname{table}{Table}{Tables}
\crefname{table}{Tab.}{Tabs.}

\usepackage{color}
\definecolor{runpei-blue}{RGB}{0, 113, 188}

\usepackage[skip=0pt]{caption} 
\begin{document}

\title{SimpleVSF: VLM-Scoring Fusion for Trajectory Prediction of End-to-End Autonomous Driving}
\author{
    Peiru Zheng,~
    Yun Zhao,~
    Zhan Gong,~
    Hong Zhu,~
    Shaohua Wu\\
    IEIT Systems\\
    {\tt\small\{zhengpeiru,zhaoyun02,gongzhan01,zhuhongbj,wushaohua\}@ieisystem.com}
}
\maketitle

\begin{abstract}
End-to-end autonomous driving has emerged as a
promising paradigm for achieving robust and intelligent driving
policies. However, existing end-to-end methods still face significant
challenges, such as suboptimal decision-making in complex scenarios. In this paper, we propose SimpleVSF (\textbf{Simple}
\textbf{V}LM-\textbf{S}coring \textbf{F}usion), a novel framework that enhances end-to-end planning
by leveraging the cognitive capabilities of Vision-Language Models
(VLMs) and advanced trajectory fusion techniques. We utilize the
conventional scorers and the novel VLM-enhanced scorers. And we leverage a robust weight fusioner for
quantitative aggregation and a powerful VLM-based fusioner for
qualitative, context-aware decision-making. As the leading approach in
the ICCV 2025 NAVSIM v2 End-to-End Driving Challenge, our SimpleVSF
framework demonstrates state-of-the-art performance, achieving a
superior balance between safety, comfort, and efficiency.
\end{abstract}

\section{Introduction}\label{introduction}

End-to-end autonomous driving has
emerged as a promising direction for achieving robust and intelligent
driving policies.\cite{RN1, RN2, RN3, RN4} Compared to the traditional
modular pipeline,\cite{RN5} this approach is designed to
minimize error accumulation and information loss between individual
components through end-to-end optimization. However, existing end-to-end
methods still face significant challenges, such as suboptimal
decision-making in complex long-tail scenarios and a lack of sufficient
trajectory diversity.\cite{RN6, RN7, RN8}

Within this context, trajectory generation and scoring have become
critical components of end-to-end planning systems. Recent works have
leveraged diffusion models\cite{RN4} to generate a wide variety
of high-quality trajectories, while dedicated trajectory scorers are
responsible for selecting the best option from these
candidates.\cite{RN2, RN9, RN10, RN11, RN12} Pioneering work such as Generalized
Trajectory Scoring (GTRS)\cite{RN13} has made notable progress
in this area. However, these methods often lack the necessary understanding of high-level
semantics and common sense crucial for complex real-world scenarios.
This limitation can lead to suboptimal performance when encountering
unseen or challenging situations.

Vision-language models (VLMs) have demonstrated increasingly powerful
image understanding and reasoning abilities.\cite{RN14, RN15, RN16, RN17, RN18, RN19, RN20} As
autonomous driving systems move towards more human-like decision-making,
the role of VLMs becomes increasingly critical.\cite{RN21, RN22}
VLMs can process not only
what is physically present (e.g., cars, lanes) but also the abstract
context and nuanced interactions. This capability allows the driving
policy to make informed, context-aware decisions.\cite{RN23, RN24, RN25, RN26, RN27, RN28, RN29, RN30}

To bridge the critical gap, we propose a novel planning framework,
Simple VLM-Scoring Fusion (SimpleVSF). Our core idea is to harness the
powerful scene understanding and reasoning capabilities of large VLMs by
incorporating their high-level semantic features throughout the
trajectory scoring and selection process. Our main contributions are summarized as
follows:

\textbf{VLM-Enhanced Scorers} We integrate high-level semantic features
from a VLM into several scorers, particularly those inspired by the GTRS
framework. This allows our scorers to move beyond raw sensor data and
comprehend deeper traffic intentions and scene common sense, leading to
more informed and robust trajectory evaluations.

\textbf{Weight-Driven and VLM-Driven Trajectory Fusion} We also
incorporate the VLM into the final weighted fusion stage for trajectory
selection. Through the VLM's high-level semantic assessment, we ensure
that the final generated trajectory is not only numerically optimal but
also semantically and ethically sound.

\textbf{A Novel End-to-End Planning Framework} The proposed SimpleVSF
framework offers an innovative solution for the ICCV 2025 Autonomous
Grand Challenge, effectively combining the generative power of diffusion
models with the high-level reasoning of VLMs. We provide comprehensive
validation of our method on the challenging NAVSIM dataset.

\hypertarget{method}{%
\section{Method}\label{method}}

\begin{figure*}[h!]
\centering
\includegraphics[width=\textwidth]{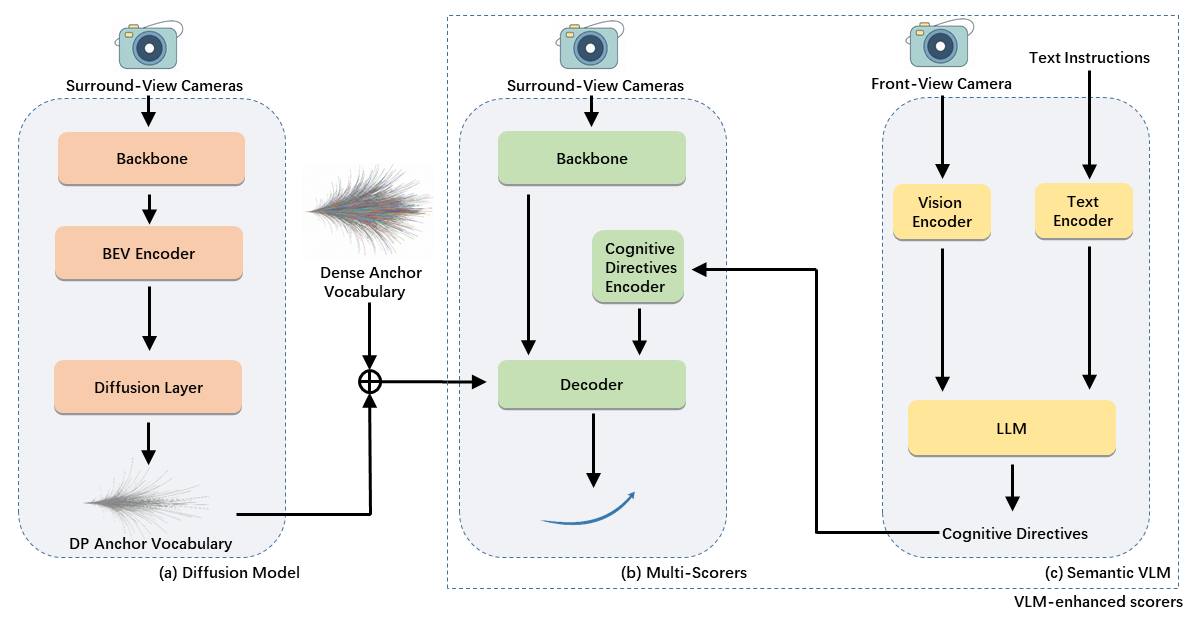}
\caption{Overall architecture of SimpleVSF.}
\label{fig:architecture}
\end{figure*}

Figure 1 illustrates the diffusion model that generates
trajectory candidates and the VLM-enhanced scorers of the SimpleVSF pipeline.

\hypertarget{trajectory-candidates-generation}{%
\subsection{Trajectory Candidates Generation}\label{trajectory-candidates-generation}}

The first stage of our pipeline is responsible for generating a diverse
set of plausible driving trajectories. We employ a diffusion-based
trajectory generator, which takes as input the ego-car's state and a
Bird's-Eye-View (BEV) representation of the surrounding
environment\cite{RN4} and generates a rich set of
candidate trajectories, i.e. anchors. Along with a super-dense vocabulary of
trajectory samples\cite{RN13}, these anchors serve as the
foundational inputs for the subsequent scoring and fusion stages.

\hypertarget{vlm-enhanced-scoring}{%
\subsection{VLM-Enhanced Scoring}\label{vlm-enhanced-scoring}}

A key innovation of our SimpleVSF framework is the hybrid approach to
trajectory scoring, which combines both perception-based
scorers with a new class of VLM-enhanced scorers. This dual approach is
designed to increase the diversity of scoring results and ensure a
comprehensive evaluation that considers both low-level geometric and
high-level semantic factors.

The scoring module takes the generated trajectory anchors and evaluates
them based on a set of criteria. These scorers are divided into two
distinct groups. The first group consists of conventional scorers based
on GTRS\cite{RN13}. The
second, more novel group of scorers is powered by a Semantic VLM module,
which provides high-level cognitive guidance. This module processes the
front-view camera images and specific text instructions to generate high-level driving directives. As shown in our framework, the
VLM is queried with a meticulously designed prompt that includes the
current state of the vehicle (speed, acceleration, and a high-level
driving command like "left" or "forward" from original NAVSIM dataset).
In response, the VLM predicts a future longitudinal directive and
lateral directive for the ego vehicle, formatted as cognitive directives
(e.g., "Accelerate, Right"). The longitudinal directive could be
keep/accelerate/decelerate/stop and the lateral directive could be
forward/left/right.

The output from the semantic VLM module is then processed by a dedicated
cognitive directives encoder. This encoder is implemented as a learnable
embedding layer, where each possible cognitive directive is mapped to a
unique vector. This process converts the VLM's abstract
linguistic instructions into a dense numerical feature, making it
compatible with the downstream scoring network. This encoded cognitive
directive is then concatenated with the ego status features and other
perceptual inputs. By feeding this enriched feature vector into the
scorers' decoders, the VLM's high-level semantic understanding is
explicitly integrated into the trajectory evaluation process.

\hypertarget{trajectory-fusion}{%
\subsection{Trajectory Fusion}\label{trajectory-fusion}}

Finally, by utilizing trajectories from multi-scorers, a trajectory fusioner selects the optimal trajectory from the candidates
proposed and scored in the previous stages. Our framework employs two
distinct fusion strategies: a weight fusioner and a
VLM-based selection fusioner.

\textbf{Weight Fusioner} The weight fusioner serves as the primary mechanism for combining scores
from multiple scorers and models to produce a single, unified score for
each candidate trajectory. This process is inspired by the aggregation
methods used in ensembles and is designed for quantitative rigor. First, the scores from individual metrics are aggregated using
a fixed-weight logarithmic sum. This initial step combines diverse
scoring aspects into a single value, with weights pre-defined to
prioritize certain critical metrics. Second, the aggregated scores are then fused using a dynamic
weighting scheme. The weights for each model's output are either
uniformly distributed or pre-assigned based on their known performance.
The final trajectory score is the sum of these weighted scores across
all models. The trajectory with the highest combined score is then
selected as the final output.

\begin{figure}[h!] 
\centering
\includegraphics[width=0.9\columnwidth]{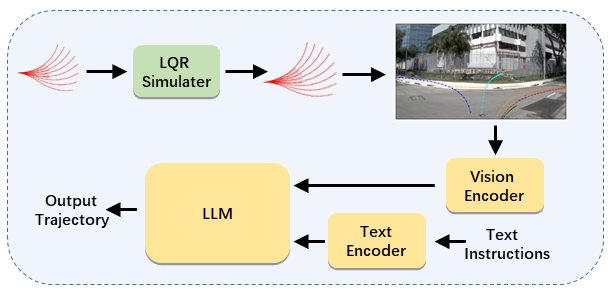} 
\caption{VLM-based trajectories selection method of VLM fusioner.}
\label{fig:vlm_fusioner}
\end{figure}

\textbf{VLM Fusioner} In addition to the quantitative fusion, our framework introduces a novel
VLM-based selection method that leverages the VLM's qualitative,
semantic reasoning for final trajectory refinement, as shown in Figure 2. First, we identify the top-ranked trajectory from each individual
scorer. These high-performing trajectories are then passed through an
LQR (Linear Quadratic Regulator) simulator\cite{RN31} to
generate smooth and kinematically feasible simulated trajectories. Next, these simulated trajectories are visualized and rendered into the
front-view camera images of the driving scene. This image is then presented to the VLM, which is
prompted to perform a final selection. The VLM's ultimate choice is then adopted as the final, planned
trajectory.

\hypertarget{experiments}{%
\section{Experiments}\label{experiments}}

\hypertarget{dataset-and-metrics}{%
\subsection{Dataset and metrics}\label{dataset-and-metrics}}

Our proposed SimpleVSF framework is evaluated on the NAVSIM dataset.\cite{RN32} We
utilize different splits of the dataset for distinct phases of our work:
the Navtrain split for training our models, the Navhard split for
ablation studies, and the Private\_test\_hard split for competition
submission.

The performance of our method is primarily measured by the Extended
Predictive Driver Model Score (EPDMS), an advanced metric introduced in
NAVSIM v2.\cite{RN33} The EPDMS is a composite metric that combines multiple weighted
subscores and multiplicative penalties to provide a comprehensive
evaluation of a planner's performance. The full composition of the EPDMS
includes: No at-fault Collisions (NC), Drivable Area Compliance (DAC), Driving Direction Compliance (DDC), Traffic
Light Compliance (TLC), Ego Progress (EP), Time to Collision (TTC), Lane Keeping (LK), History Comfort (HC), Extended Comfort (EC).

\hypertarget{implementation-details}{%
\subsection{Implementation Details}\label{implementation-details}}

All models in our framework were trained on a cluster of 8 NVIDIA A800
GPUs.

For the diffusion model, we directly adopted the pre-trained model
weights from the GTRS framework\cite{RN13} to generate the
initial set of diverse candidate trajectories.

The scorers, which are responsible for evaluating these trajectories,
were trained for 20 epochs with a global batch size of 64 (8 GPUs x 8
samples per GPU). We used a learning rate of 2×10\textsuperscript{-4}
for optimization.

The semantic VLM module, based on Qwen2VL-2B,\cite{RN34} was
fine-tuned on the Navtrain split of the NAVSIM dataset to generate the
cognitive directives. The fine-tuning process was carried out over 4
epochs with a global batch size of 32 (4 samples per GPU), and a
learning rate of 2×10\textsuperscript{-5}.

For the VLM fusioner, we utilized a larger and more capable model,
Qwen2.5VL-72B\cite{RN16}. This model was not fine-tuned.
Instead, we employed a few-shot prompting strategy to directly perform
trajectory selection at inference time, leveraging its powerful
out-of-the-box reasoning and visual-language capabilities.

\hypertarget{main-results}{%
\subsection{Main Results}\label{main-results}}

\begin{table*}[h!]
\centering
\caption{Performance on Private\_test\_hard split, i.e. ICCV 2025 NAVSIM v2 End-to-End Driving Challenge.}
\label{tab:performance}
\begin{tabular}{@{}llllllllllll@{}}
\toprule
\textbf{Method/Team} & \textbf{Stage} & \textbf{NC} & \textbf{DAC} & \textbf{DDC} & \textbf{TLC} & \textbf{EP} & \textbf{TTC} & \textbf{LK} & \textbf{HC} & \textbf{EC} & \textbf{EPDMS} \\
\midrule
\multirow{2}{*}{bjtu\_jia\_team\&qcraft} & Stage I & \textbf{98.21} & \textbf{100} & 99.64 & \textbf{100} & 80.84 & 98.57 & 90.00 & 94.29 & 57.14 & \multirow{2}{*}{51.31} \\
& Stage II & 88.90 & 95.44 & 97.92 & 96.84 & 77.92 & 88.02 & 56.62 & 98.31 & 64.43 & \\
\midrule
\multirow{2}{*}{DRL\_CASIA\&XIAOMI} & Stage I & 96.43 & 99.29 & \textbf{100} & 98.57 & \textbf{85.63} & \textbf{99.29} & 93.57 & \textbf{95.00} & \textbf{70.00} & \multirow{2}{*}{51.08} \\
& Stage II & 87.51 & \textbf{96.59} & 97.04 & 96.63 & \textbf{84.21} & 86.30 & 55.41 & 98.91 & \textbf{74.74} & \\
\midrule
\multirow{2}{*}{DiffVLA++} & Stage I & \textbf{98.21} & 98.57 & \textbf{100} & 99.29 & 79.51 & 98.57 & 95.00 & 92.86 & 50.00 & \multirow{2}{*}{49.12} \\
& Stage II & 88.77 & 95.32 & 97.22 & \textbf{98.17} & 73.43 & 87.99 & \textbf{59.45} & \textbf{98.98} & 52.98 & \\
\midrule
\multirow{2}{*}{SimpleVSF (Our)} & Stage I & \textbf{98.21} & 99.29 & 99.29 & \textbf{100} & 81.30 & 98.57 & \textbf{95.71} & 93.57 & 51.43 & \multirow{2}{*}{\textbf{53.06}} \\
& Stage II & \textbf{91.20} & 95.40 & \textbf{98.77} & 97.11 & 79.98 & \textbf{88.69} & 56.15 & 97.43 & 56.82 & \\
\bottomrule
\end{tabular}
\end{table*}

Table 1 presents the final EPDMS scores and their corresponding
sub-metrics for the top-performing teams on the Private\_test\_hard
split of the ICCV 2025 NAVSIM v2 End-to-End Driving Challenge. Our
proposed SimpleVSF framework achieved first place on the leaderboard.

As shown in Table 1, our method achieved an overall EPDMS score of
53.06. For Stage I, it scored a
perfect 100 on TLC, a near-perfect 99.29 on DAC and DDC, showcasing the
model's robustness and its ability to adhere to critical traffic rules.
For both Stage I and Stage II, our NC scores take a leading position
among all competitors. While other methods may excel
in certain aspects, our SimpleVSF achieves a superior balance across a
wide range of metrics.

\hypertarget{ablation-study}{%
\subsection{Ablation Study}\label{ablation-study}}

To systematically evaluate the contribution of each component within our
SimpleVSF framework, we conducted a comprehensive ablation study on the
Navhard split. The results are presented in Table 2.

\begin{table}[h!] 
\centering
\caption{Results of SimpleVSF of different settings on Navhard split. Version A: No-VLM scorer with V2-99 backbone; Version B: No-VLM scorer with EVA-L backbone; Version C: No-VLM scorer with ViT-L backbone; Version D: VLM-enhanced scorer with V2-99 backbone; Version E: VLM-enhanced scorer with ViT-L backbone. WF: Weight Fusioner; VLMF: VLM Fusioner.}
\label{tab:results_simplevsf}
\begin{tabular}{@{}p{2.5cm}ccc@{}}
\toprule
\textbf{Method} & \textbf{EPDMS I} & \textbf{EPDMS II} & \textbf{EPDMS} \\
\midrule
Version A & 73.00 & 57.28 & 42.51 \\
Version B & 74.85 & 57.87 & 43.61 \\
Version C & 73.89 & 60.03 & 45.41 \\
Version D & 75.33 & 56.28 & 43.30 \\
Version E & 72.52 & 59.56 & 43.66 \\
\textbf{WF B+C+D+E} & \textbf{75.37} & \textbf{61.90} & \textbf{47.18} \\
\emph{VLMF A+B+C} & \emph{74.82} & \emph{62.58} & \emph{47.68} \\
\bottomrule
\end{tabular}
\end{table}

\textbf{Impact of Different Backbones} We utilize three distinct backbones, i.e. V2-99\cite{RN36},
EVA-L\cite{RN37}, ViT-L\cite{RN38, RN39}. Versions A through
F of our models, which use GTRS scorers,\cite{RN13} serve as
our baselines. The results show that the choice of backbone plays a
significant role in performance. The ViT-L backbone consistently
outperformed the others.

\textbf{Effectiveness of VLM-Enhanced Scorers} Versions D and E, which incorporate our VLM-enhanced scorers,
demonstrate the value of semantic guidance. While their individual
performance is comparable to or slightly lower than the best-performing
traditional scorers (e.g., Version C), the real strength of the
VLM-enhanced scorers is in their potential for fusion.

\textbf{Performance of Trajectory Fusion Strategies} The most significant performance gains were observed through our fusion
strategies. The WF B+C+D+E achieved a high score of 47.18 on the Navhard split. Ultimately, we also used the fusion results of these four scorers on the Private\_test\_hard split. The VLMF A+B+C, also achieved an impressive EPDMS of 47.68, but we did not use this fusion strategy for leaderboard submission.

\hypertarget{conclusion}{%
\section{Conclusion}\label{conclusion}}

In this paper, we have presented SimpleVSF, a novel and effective
framework that bridges the gap between traditional trajectory planning
and the semantic understanding offered by Vision-Language Models. By
integrating a diffusion-based trajectory generator, a diverse set of
scorers, and two fusioners, our approach addresses key limitations of
existing end-to-end driving systems.

{\small
\bibliographystyle{unsrt}
\bibliography{ref.bib} 

\begin{thebibliography}{10}

\bibitem{RN1}
Wenchao Sun, Xuewu Lin, Yining Shi, Chuang Zhang, Haoran Wu, and Sifa Zheng.
\newblock Sparsedrive: End-to-end autonomous driving via sparse scene representation.
\newblock {\em 2025 IEEE International Conference on Robotics and Automation (ICRA)}, pages 8795--8801, 2025.

\bibitem{RN2}
Bencheng Liao, Shaoyu Chen, Haoran Yin, Bo~Jiang, Cheng Wang, Sixu Yan, Xinbang Zhang, Xiangyu Li, Ying Zhang, Qian Zhang, et~al.
\newblock Diffusiondrive: Truncated diffusion model for end-to-end autonomous driving.
\newblock {\em Proceedings of the Computer Vision and Pattern Recognition Conference}, pages 12037--12047, 2025.

\bibitem{RN3}
Xiaosong Jia, Junqi You, Zhiyuan Zhang, and Junchi Yan.
\newblock Drivetransformer: Unified transformer for scalable end-to-end autonomous driving.
\newblock {\em arXiv preprint arXiv:2503.07656}, 2025.

\bibitem{RN4}
Ekim Yurtsever, Jacob Lambert, Alexander Carballo, and Kazuya Takeda.
\newblock A survey of autonomous driving: Common practices and emerging technologies.
\newblock {\em IEEE access}, 8:58443--58469, 2020.

\bibitem{RN5}
Kashyap Chitta, Aditya Prakash, Bernhard Jaeger, Zehao Yu, Katrin Renz, and Andreas Geiger.
\newblock Transfuser: Imitation with transformer-based sensor fusion for autonomous driving.
\newblock {\em IEEE transactions on pattern analysis and machine intelligence}, 45(11):12878--12895, 2022.

\bibitem{RN6}
Yihan Hu, Jiazhi Yang, Li~Chen, Keyu Li, Chonghao Sima, Xizhou Zhu, Siqi Chai, Senyao Du, Tianwei Lin, Wenhai Wang, et~al.
\newblock Planning-oriented autonomous driving.
\newblock {\em Proceedings of the IEEE/CVF conference on computer vision and pattern recognition}, pages 17853--17862, 2023.

\bibitem{RN7}
Bo~Jiang, Shaoyu Chen, Qing Xu, Bencheng Liao, Jiajie Chen, Helong Zhou, Qian Zhang, Wenyu Liu, Chang Huang, and Xinggang Wang.
\newblock Vad: Vectorized scene representation for efficient autonomous driving.
\newblock {\em Proceedings of the IEEE/CVF International Conference on Computer Vision}, pages 8340--8350, 2023.

\bibitem{RN8}
Cheng Chi, Zhenjia Xu, Siyuan Feng, Eric Cousineau, Yilun Du, Benjamin Burchfiel, Russ Tedrake, and Shuran Song.
\newblock Diffusion policy: Visuomotor policy learning via action diffusion.
\newblock {\em The International Journal of Robotics Research}, page 02783649241273668, 2023.

\bibitem{RN9}
Sourav Biswas, Sergio Casas, Quinlan Sykora, Ben Agro, Abbas Sadat, and Raquel Urtasun.
\newblock Quad: Query-based interpretable neural motion planning for autonomous driving.
\newblock {\em 2024 IEEE International Conference on Robotics and Automation (ICRA)}, pages 14236--14243, 2024.

\bibitem{RN10}
Shaoyu Chen, Bo~Jiang, Hao Gao, Bencheng Liao, Qing Xu, Qian Zhang, Chang Huang, Wenyu Liu, and Xinggang Wang.
\newblock Vadv2: End-to-end vectorized autonomous driving via probabilistic planning.
\newblock {\em arXiv preprint arXiv:2402.13243}, 2024.

\bibitem{RN11}
Kailin Li, Zhenxin Li, Shiyi Lan, Yuan Xie, Zhizhong Zhang, Jiayi Liu, Zuxuan Wu, Zhiding Yu, and Jose~M Alvarez.
\newblock Hydra-mdp++: Advancing end-to-end driving via expert-guided hydra-distillation.
\newblock {\em arXiv preprint arXiv:2503.12820}, 2025.

\bibitem{RN12}
Chonghao Sima, Kashyap Chitta, Zhiding Yu, Shiyi Lan, Ping Luo, Andreas Geiger, Hongyang Li, and Jose~M Alvarez.
\newblock Centaur: Robust end-to-end autonomous driving with test-time training.
\newblock {\em arXiv preprint arXiv:2503.11650}, 2025.

\bibitem{RN13}
Zhenxin Li, Wenhao Yao, Zi~Wang, Xinglong Sun, Joshua Chen, Nadine Chang, Maying Shen, Zuxuan Wu, Shiyi Lan, and Jose~M Alvarez.
\newblock Generalized trajectory scoring for end-to-end multimodal planning.
\newblock {\em arXiv preprint arXiv:2506.06664}, 2025.

\bibitem{RN14}
Haotian Liu, Chunyuan Li, Qingyang Wu, and Yong~Jae Lee.
\newblock Visual instruction tuning.
\newblock {\em Advances in neural information processing systems}, 36:34892--34916, 2023.

\bibitem{RN15}
Haotian Liu, Chunyuan Li, Yuheng Li, and Yong~Jae Lee.
\newblock Improved baselines with visual instruction tuning.
\newblock {\em Proceedings of the IEEE/CVF conference on computer vision and pattern recognition}, pages 26296--26306, 2024.

\bibitem{RN16}
Shuai Bai, Keqin Chen, Xuejing Liu, Jialin Wang, Wenbin Ge, Sibo Song, Kai Dang, Peng Wang, Shijie Wang, Jun Tang, et~al.
\newblock Qwen2. 5-vl technical report.
\newblock {\em arXiv preprint arXiv:2502.13923}, 2025.

\bibitem{RN17}
Jinguo Zhu, W~Wang, Z~Chen, Z~Liu, S~Ye, L~Gu, H~Tian, Y~Duan, W~Su, J~Shao, et~al.
\newblock Internvl3: Exploring advanced training and test-time recipes for open-source multimodal models.
\newblock {\em arXiv preprint arXiv:2504.10479}, 9, 2025.

\bibitem{RN18}
Weihan Wang, Qingsong Lv, Wenmeng Yu, Wenyi Hong, Ji~Qi, Yan Wang, Junhui Ji, Zhuoyi Yang, Lei Zhao, Song XiXuan, et~al.
\newblock Cogvlm: Visual expert for pretrained language models.
\newblock {\em Advances in Neural Information Processing Systems}, 37:121475--121499, 2024.

\bibitem{RN19}
Wenyi Hong, Weihan Wang, Ming Ding, Wenmeng Yu, Qingsong Lv, Yan Wang, Yean Cheng, Shiyu Huang, Junhui Ji, Zhao Xue, et~al.
\newblock Cogvlm2: Visual language models for image and video understanding.
\newblock {\em arXiv preprint arXiv:2408.16500}, 2024.

\bibitem{RN20}
Kimi Team, Angang Du, Bohong Yin, Bowei Xing, Bowen Qu, Bowen Wang, Cheng Chen, Chenlin Zhang, Chenzhuang Du, Chu Wei, et~al.
\newblock Kimi-vl technical report.
\newblock {\em arXiv preprint arXiv:2504.07491}, 2025.

\bibitem{RN21}
Tsun-Hsuan Wang, Alaa Maalouf, Wei Xiao, Yutong Ban, Alexander Amini, Guy Rosman, Sertac Karaman, and Daniela Rus.
\newblock Drive anywhere: Generalizable end-to-end autonomous driving with multi-modal foundation models.
\newblock In {\em 2024 IEEE International Conference on Robotics and Automation (ICRA)}, pages 6687--6694. IEEE, 2024.

\bibitem{RN22}
Li~Chen, Penghao Wu, Kashyap Chitta, Bernhard Jaeger, Andreas Geiger, and Hongyang Li.
\newblock End-to-end autonomous driving: Challenges and frontiers.
\newblock {\em IEEE Transactions on Pattern Analysis and Machine Intelligence}, 2024.

\bibitem{RN23}
Junwei You, Haotian Shi, Zhuoyu Jiang, Zilin Huang, Rui Gan, Keshu Wu, Xi~Cheng, Xiaopeng Li, and Bin Ran.
\newblock V2x-vlm: End-to-end v2x cooperative autonomous driving through large vision-language models.
\newblock {\em arXiv preprint arXiv:2408.09251}, 2024.

\bibitem{RN24}
Yi~Xu, Yuxin Hu, Zaiwei Zhang, Gregory~P Meyer, Siva~Karthik Mustikovela, Siddhartha Srinivasa, Eric~M Wolff, and Xin Huang.
\newblock Vlm-ad: End-to-end autonomous driving through vision-language model supervision.
\newblock {\em arXiv preprint arXiv:2412.14446}, 2024.

\bibitem{RN25}
Ziang Guo, Zakhar Yagudin, Artem Lykov, Mikhail Konenkov, and Dzmitry Tsetserukou.
\newblock Vlm-auto: Vlm-based autonomous driving assistant with human-like behavior and understanding for complex road scenes.
\newblock {\em 2024 2nd International Conference on Foundation and Large Language Models (FLLM)}, pages 501--507, 2024.

\bibitem{RN26}
Zilin Huang, Zihao Sheng, Yansong Qu, Junwei You, and Sikai Chen.
\newblock Vlm-rl: A unified vision language models and reinforcement learning framework for safe autonomous driving.
\newblock {\em Transportation Research Part C: Emerging Technologies}, 180:105321, 2025.

\bibitem{RN27}
Pei Liu, Haipeng Liu, Haichao Liu, Xin Liu, Jinxin Ni, and Jun Ma.
\newblock Vlm-e2e: Enhancing end-to-end autonomous driving with multimodal driver attention fusion.
\newblock {\em arXiv preprint arXiv:2502.18042}, 2025.

\bibitem{RN28}
Chonghao Sima, Katrin Renz, Kashyap Chitta, Li~Chen, Hanxue Zhang, Chengen Xie, Jens Bei{\ss}wenger, Ping Luo, Andreas Geiger, and Hongyang Li.
\newblock Drivelm: Driving with graph visual question answering.
\newblock {\em European conference on computer vision}, pages 256--274, 2024.

\bibitem{RN29}
Bo~Jiang, Shaoyu Chen, Bencheng Liao, Xingyu Zhang, Wei Yin, Qian Zhang, Chang Huang, Wenyu Liu, and Xinggang Wang.
\newblock Senna: Bridging large vision-language models and end-to-end autonomous driving.
\newblock {\em arXiv preprint arXiv:2410.22313}, 2024.

\bibitem{RN30}
Peiru Zheng, Yun Zhao, Zhan Gong, Hong Zhu, and Shaohua Wu.
\newblock Simplellm4ad: An end-to-end vision-language model with graph visual question answering for autonomous driving.
\newblock {\em arXiv preprint arXiv:2407.21293}, 2024.

\bibitem{RN31}
Norman Lehtomaki, NJAM Sandell, and Michael Athans.
\newblock Robustness results in linear-quadratic gaussian based multivariable control designs.
\newblock {\em IEEE Transactions on Automatic Control}, 26(1):75--93, 2003.

\bibitem{RN32}
Daniel Dauner, Marcel Hallgarten, Tianyu Li, Xinshuo Weng, Zhiyu Huang, Zetong Yang, Hongyang Li, Igor Gilitschenski, Boris Ivanovic, Marco Pavone, et~al.
\newblock Navsim: Data-driven non-reactive autonomous vehicle simulation and benchmarking.
\newblock {\em Advances in Neural Information Processing Systems}, 37:28706--28719, 2024.

\bibitem{RN33}
Wei Cao, Marcel Hallgarten, Tianyu Li, Daniel Dauner, Xunjiang Gu, Caojun Wang, Yakov Miron, Marco Aiello, Hongyang Li, Igor Gilitschenski, et~al.
\newblock Pseudo-simulation for autonomous driving.
\newblock {\em arXiv preprint arXiv:2506.04218}, 2025.

\bibitem{RN34}
Peng Wang, Shuai Bai, Sinan Tan, Shijie Wang, Zhihao Fan, Jinze Bai, Keqin Chen, Xuejing Liu, Jialin Wang, Wenbin Ge, et~al.
\newblock Qwen2-vl: Enhancing vision-language model's perception of the world at any resolution.
\newblock {\em arXiv preprint arXiv:2409.12191}, 2024.

\bibitem{RN36}
Youngwan Lee, Joong-won Hwang, Sangrok Lee, Yuseok Bae, and Jongyoul Park.
\newblock An energy and gpu-computation efficient backbone network for real-time object detection.
\newblock {\em Proceedings of the IEEE/CVF conference on computer vision and pattern recognition workshops}, pages 0--0, 2019.

\bibitem{RN37}
Yuxin Fang, Quan Sun, Xinggang Wang, Tiejun Huang, Xinlong Wang, and Yue Cao.
\newblock Eva-02: A visual representation for neon genesis.
\newblock {\em Image and Vision Computing}, 149:105171, 2024.

\bibitem{RN38}
Alexey Dosovitskiy, Lucas Beyer, Alexander Kolesnikov, Dirk Weissenborn, Xiaohua Zhai, Thomas Unterthiner, Mostafa Dehghani, Matthias Minderer, Georg Heigold, Sylvain Gelly, et~al.
\newblock An image is worth 16x16 words: Transformers for image recognition at scale.
\newblock {\em arXiv preprint arXiv:2010.11929}, 2020.

\bibitem{RN39}
Lihe Yang, Bingyi Kang, Zilong Huang, Xiaogang Xu, Jiashi Feng, and Hengshuang Zhao.
\newblock Depth anything: Unleashing the power of large-scale unlabeled data.
\newblock {\em Proceedings of the IEEE/CVF conference on computer vision and pattern recognition}, pages 10371--10381, 2024.

\end{thebibliography}
}

\end{document}